\documentclass[11pt,a4paper]{article}
\usepackage[hyperref]{acl2020}
\usepackage{times}
\usepackage{latexsym}

\usepackage{microtype}

\usepackage{hhline}
\usepackage{graphicx}
\usepackage{url}
\usepackage{multirow}
\usepackage{xcolor}
\definecolor{gainsboro}{rgb}{0.86, 0.86, 0.86}
\usepackage{stfloats}

\aclfinalcopy 
\def\aclpaperid{2092} 

\usepackage{fancyhdr}
\usepackage{geometry}
\usepackage{layout}
\title{Unsupervised Domain Clusters in Pretrained Language Models}

\author{Roee Aharoni$^{1}$ \& Yoav Goldberg$^{1,2}$ \\
  $^1$ Computer Science Department, Bar Ilan University \\
  $^2$ Allen Institute for Artificial Intelligence \\
  {\tt first.last@gmail.com}}
  
\date{}

\begin{document}
\maketitle
\begin{abstract}
The notion of ``in-domain data'' in NLP is often over-simplistic and vague, as textual data varies in many nuanced linguistic aspects such as topic, style or level of formality. In addition, domain labels are many times unavailable, making it challenging to build domain-specific systems.  
We show that massive pre-trained language models implicitly learn sentence representations that cluster by domains without supervision -- suggesting a simple data-driven definition of domains in textual data. We harness this property and propose domain data selection methods based on such models, which require only a small set of in-domain monolingual data. We evaluate our data selection methods for neural machine translation across five diverse domains, where they outperform an established approach as measured by both BLEU and by precision and recall of sentence selection with respect to an oracle.
\end{abstract}

\thispagestyle{fancy}

\section{Introduction}


It is common knowledge in modern NLP that using large amounts of high-quality training data is a key aspect in building successful machine-learning based systems. For this reason, a major challenge when building such systems is obtaining data in the domain of interest. But what defines a domain? Natural language varies greatly across topics, styles, levels of formality, genres and many other linguistic nuances \citep{van-der-wees-etal-2015-whats,van-der-wees-thesis-whats,niu-etal-2017-study}. This overwhelming diversity of language makes it hard to find the right data for the task, as it is nearly impossible to well-define the exact requirements from such data with respect to all the aforementioned aspects. On top of that, domain labels are usually unavailable -- e.g. in large-scale web-crawled data like Common Crawl\footnote{\url{https://commoncrawl.org/}} which was recently used to train state-of-the-art pretrained language models for various tasks \cite{raffel2019exploring}.


\begin{figure}[t]
    \centering
    \fcolorbox{gainsboro}{white}{\includegraphics[scale=0.35,trim={1.5cm 0.5cm 0.5cm 1.5cm},clip]{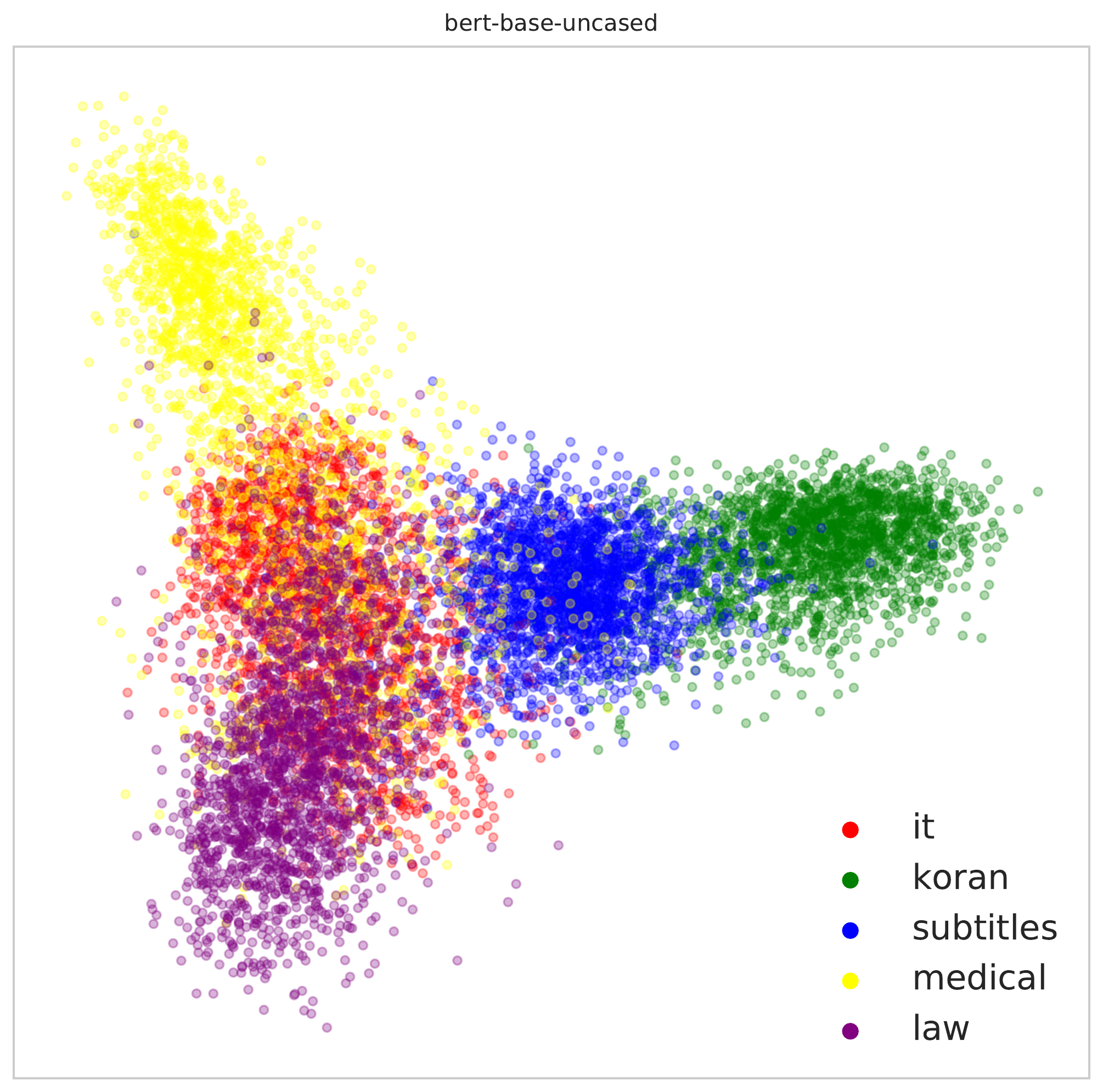}}
    \caption{A 2D visualization of average-pooled BERT hidden-state sentence representations using PCA. The colors represent the domain for each sentence.}
    \label{fig:bert_domains}
    \vspace{-15px}
\end{figure}



\emph{Domain data selection} is the task of selecting the most appropriate data for a domain from a large corpus given a smaller set of in-domain data \cite{moore-lewis-2010-intelligent, axelrod-etal-2011-domain,duh-etal-2013-adaptation,silva-etal-2018-extracting}. In this work, we propose to use the recent, highly successful self-supervised pre-trained language models, e.g. \newcite{devlin-etal-2019-bert,liu2019roberta} for domain data selection. As pretrained LMs demonstrate state-of-the-art performance across many NLP tasks after being trained on massive amounts of data, we hypothesize that the robust representations they learn can be useful for mapping sentences to domains in an unsupervised, data-driven approach. We show that these models indeed learn to cluster sentence representations to domains without further supervision (e.g. Figure \ref{fig:bert_domains}), and quantify this phenomenon by fitting Gaussian Mixture Models (GMMs) to the learned representations and measuring the purity of the resulting unsupervised clustering. We then propose methods to leverage these emergent domain clusters for domain data selection in two ways: 
\begin{itemize}
    \item Via distance-based retrieval in the sentence embedding space induced by the pretrained language model.
    \item By fine-tuning the pretrained language model for binary classification, where positive examples are from the domain of interest.
\end{itemize}

Our methods enable to select relevant data for the task while requiring only a small set of monolingual in-domain data. As they are based solely on the representations learned by self-supervised LMs, they do not require additional domain labels which are usually vague and over-simplify the notion of domain in textual data.
We evaluate our method on data selection for neural machine translation (NMT) using the multi-domain German-English parallel corpus composed by \citet{koehn-knowles-2017-six}. Our data selection methods enable to train NMT models that outperform those trained using the well-established cross-entropy difference method of \citet{moore-lewis-2010-intelligent} across five diverse domains, achieving a recall of more than 95\% in all cases with respect to an oracle that selects the ``true'' in-domain data.

Our contributions in this work are as follows. First, we show that pre-trained language models are highly capable of clustering textual data to domains with high accuracy in a purely unsupervised manner.
Second, we propose methods to select in-domain data based on this property using vector-space retrieval and positive-unlabeled fine-tuning of pretrained language models for binary classification. Third, we show the applicability of our proposed data selection methods on a popular benchmark for domain adaptation in machine translation. An additional contribution is a new, improved data split we create for this benchmark, as we point on issues with previous splits used in the literature. The code and data for this work is publicly available.\footnote{\url{https://github.com/roeeaharoni/unsupervised-domain-clusters}} We hope this work will encourage more research on understanding the data landscape in NLP, enabling to ``find the right data for the task'' in the age of massive models and diverse data sources.



\section{Emerging Domain Clusters in Pretrained Language Models}
\label{sec:clusters}
\subsection{Motivation}
The proliferation of massive pretrained neural language models such as ELMo \cite{peters-etal-2018-deep}, BERT \cite{devlin-etal-2019-bert} or RoBERTa \cite{liu2019roberta} has enabled great progress on many NLP benchmarks \cite{wang-etal-2018-glue, wang2019superglue}. Larger and larger models trained on billions of tokens of raw text are released in an ever-increasing pace \cite{raffel2019exploring}, enabling the NLP community to fine-tune them for the task of interest. While many works tried to ``probe'' those models for the morphological, syntactic and semantic information they capture \cite{tenney-etal-2019-bert,goldberg2019assessing,clark2019does}, an important aspect of language remained overlooked in this context -- the \textit{domain} the data comes from, often referred to as the ``data distribution''.

The definition of domain is many times vague and over-simplistic (e.g. ``medical text'' may be used for biomedical research papers and for clinical conversations between doctors and patients, although the two vary greatly in topic, formality etc.). A common definition treats a domain as a data source: ``a domain is defined by a corpus from a specific source, and may differ from other domains in topic, genre, style, level of formality, etc.'' \cite{koehn-knowles-2017-six}. We claim that a more data-driven definition should take place, as different data sources may have sentences with similar traits and vice versa - a single massive web-crawled corpus contains texts in numerous styles, topics and registers. Our analysis in Section \ref{sec:clusters} shows examples for such cases, e.g. a sentence discussing ``Viruses and virus-like organisms'' in a legal corpus. 


\begin{table*}[!t]
\begin{center}
\begin{tabular}{lllllll}
\cline{2-4} 
\multicolumn{1}{c|}{}               & \multicolumn{1}{c|}{k=5} & \multicolumn{1}{c|}{k=10} & \multicolumn{1}{c|}{k=15} \\ \cline{1-4} 
\multicolumn{1}{|l|}{Random}      & \multicolumn{1}{c|}{15.08 ($\pm$0.0)}    & \multicolumn{1}{c|}{16.77 ($\pm$0.0)}     & \multicolumn{1}{c|}{17.78 ($\pm$0.0)}  \\ \cline{1-4} 
\multicolumn{1}{|l|}{LDA}      & \multicolumn{1}{c|}{24.31 ($\pm$0.99)}    & \multicolumn{1}{c|}{26.73 ($\pm$2.19)}     & \multicolumn{1}{c|}{30.79 ($\pm$2.97)}    \\ \cline{1-4} 
\\ 
\end{tabular}
\vspace{-10px}
\begin{tabular}{lllllll}
 & \multicolumn{3}{c}{with PCA (n=50)}                                             & \multicolumn{3}{c}{without PCA}                  \\ \cline{2-7} 
 \multicolumn{1}{c|}{}               & \multicolumn{1}{c|}{k=5} & \multicolumn{1}{c|}{k=10} & \multicolumn{1}{c|}{k=15} & \multicolumn{1}{c|}{k=5} & \multicolumn{1}{c|}{k=10} & \multicolumn{1}{c|}{k=15} \\ \hline
\multicolumn{1}{|l|}{word2vec}      & \multicolumn{1}{c|}{53.65 ($\pm$0.79)}    & \multicolumn{1}{c|}{68.14 ($\pm$2.58)}     & \multicolumn{1}{c|}{73.44 ($\pm$0.68)}     & \multicolumn{1}{c|}{45.93}    & \multicolumn{1}{c|}{65.80}     & \multicolumn{1}{c|}{76.26}     \\ \hline
\multicolumn{1}{|l|}{BERT-base}     & \multicolumn{1}{c|}{\textbf{87.66 ($\pm$0.24)}}    & \multicolumn{1}{c|}{88.02 ($\pm$1.10)}     & \multicolumn{1}{c|}{88.37 ($\pm$0.66)}     & \multicolumn{1}{c|}{\textbf{85.74}}    & \multicolumn{1}{c|}{85.08}     & \multicolumn{1}{c|}{86.37}     \\ \hline
\multicolumn{1}{|l|}{BERT-large}    & \multicolumn{1}{c|}{85.64 ($\pm$6.13)}    & \multicolumn{1}{c|}{87.61 ($\pm$0.26)}     & \multicolumn{1}{c|}{89.07 ($\pm$0.53)}     & \multicolumn{1}{c|}{68.56}    & \multicolumn{1}{c|}{\textbf{86.53}}     & \multicolumn{1}{c|}{86.99}     \\ \hline
\multicolumn{1}{|l|}{DistillBERT}  & \multicolumn{1}{c|}{83.68 ($\pm$7.14)}    & \multicolumn{1}{c|}{86.31 ($\pm$0.86)}     & \multicolumn{1}{c|}{87.53 ($\pm$0.85)}     & \multicolumn{1}{c|}{79.00}    & \multicolumn{1}{c|}{86.42}     & \multicolumn{1}{c|}{\textbf{88.14}}     \\ \hline
\multicolumn{1}{|l|}{RoBERTa-base}  & \multicolumn{1}{c|}{79.05 ($\pm$0.10)}    & \multicolumn{1}{c|}{86.39 ($\pm$0.90)}     & \multicolumn{1}{c|}{86.51 ($\pm$0.28)}     & \multicolumn{1}{c|}{70.21}    & \multicolumn{1}{c|}{80.35}     & \multicolumn{1}{c|}{81.49}     \\ \hline
\multicolumn{1}{|l|}{RoBERTa-large} & \multicolumn{1}{c|}{80.61 ($\pm$0.33)}    & \multicolumn{1}{c|}{\textbf{89.04 ($\pm$0.15)}}     & \multicolumn{1}{c|}{\textbf{89.94 ($\pm$0.23)}}     & \multicolumn{1}{c|}{69.88}    & \multicolumn{1}{c|}{81.07}     & \multicolumn{1}{c|}{85.91}     \\ \hline
\multicolumn{1}{|l|}{GPT-2}          & \multicolumn{1}{c|}{70.30 ($\pm$0.05)}    & \multicolumn{1}{c|}{84.76 ($\pm$0.30)}     & \multicolumn{1}{c|}{82.56 ($\pm$1.29)}     & \multicolumn{1}{c|}{37.82}    & \multicolumn{1}{c|}{39.02}     & \multicolumn{1}{c|}{41.45}     \\ \hline
\multicolumn{1}{|l|}{XLNet}         & \multicolumn{1}{c|}{55.72 ($\pm$0.69)}    & \multicolumn{1}{c|}{68.17 ($\pm$3.93)}     & \multicolumn{1}{c|}{72.65 ($\pm$1.92)}     & \multicolumn{1}{c|}{30.36}    & \multicolumn{1}{c|}{32.96}     & \multicolumn{1}{c|}{48.55}     \\ \hline

\end{tabular}
\end{center}
\caption{Unsupervised domain clustering as measured by purity for the different models. Best results are marked in bold for each setting.}
\label{tab:cross_domain_accuracy}
\vspace{-10px}
\end{table*}

We hypothesize that massive pretrained LMs can learn representations that cluster to domains, as texts from similar domains will appear in similar contexts. 
We test this hypothesis across several large, publicly-available pretrained LMs; we explore both masked-language-models (MLMs) and auto-regressive LMs. 

\subsection{Method}
We encode multi-domain data at the sentence level into vector representations. We then cluster these vector representations for each model using a Gaussian Mixture Model (GMM) with $k$ pre-defined clusters. We chose GMM as our clustering approach as it allows soft assignments (vs. hard assignments as in e.g. K-means) which we think fits the task better (as a sentence can be seen as drawn from a mixture of several domain).\footnote{See further discussion comparing GMMs and K-means in \citet{daume_2009}.} In all cases, to create a sentence representation we perform average pooling of the last hidden state (before the softmax layer) for each token in the sentence.\footnote{Using the penultimate layer or others may result in better performance; we leave this for future work.} To accelerate the clustering process and enable visualization we also experiment with performing dimensionality reduction with PCA over the sentence vectors before clustering them. We experiment with k in 5, 10 and 15 to test how adding flexibility would improve the domain clustering accuracy.

\subsection{Models and Baselines}
For MLM-based models we use BERT \cite{devlin-etal-2019-bert}, DistilBERT \cite{sanh2019distilbert} and RoBERTa \cite{liu2019roberta} (in both the base and large versions). For autoregressive models we use GPT-2 \cite{radford2018improving} and XLNet \cite{yang2019xlnet}. In all cases we use the implementations from the HuggingFace Transformers toolkit \cite{Wolf2019HuggingFacesTS}. We also evaluated three additional, simpler baselines. The first is using representations from word2vec \cite{mikolov2013distributed}, where we average-pooled the word vectors for the tokens that were present in the model vocabulary. The second is using Latent Dirichlet Allocation (LDA, \citealp{blei2003latent}), which is a classic approach to unsupervised clustering of text.\footnote{We used the LDA implementation provided in the Gensim toolkit: \url{https://radimrehurek.com/gensim/}} We also report results for a baseline which assigns sentences by sampling randomly from a uniform distribution over the clusters.

\subsection{Evaluation}
To evaluate the unsupervised domain clustering we used the multi-domain corpus proposed by \citet{koehn-knowles-2017-six} which includes textual data in five diverse domains: subtitles\footnote{From \url{http://www.opensubtitles.org/}}, medical text (PDF documents from the European Medicines Agency), legal text (legislative text of the European Union), translations of the Koran, and IT-related text (manuals and localization files of open-source software). This dataset includes parallel sentences in English and German; for this experiment we used the English portion of the data. See more details on the dataset in Section \ref{sec:dataset}. We used 2000 distinct sentences from each domain. To evaluate whether the resulting clusters indeed capture the domains the data was drawn from we measure the clustering purity, which is a well-known metric for evaluating clustering \cite{manning2008introduction}. To measure the clustering purity, we assign each unsupervised cluster with the most common ``true'' domain in the sentences assigned to that cluster, and then compute the accuracy according to this majority-based cluster-domain assignment (note that in this case several unsupervised clusters can be assigned to the same domain). In cases where randomness is involved we run each experiment five times with different initializations and report the mean and variance of the purity metric for each model.

\begin{figure}[!b]
    \vspace{-15px}
    \centering
    \fcolorbox{gainsboro}{white}{\includegraphics[scale=0.36,trim={2.5cm 0.5cm 0.5cm 1.5cm},clip]{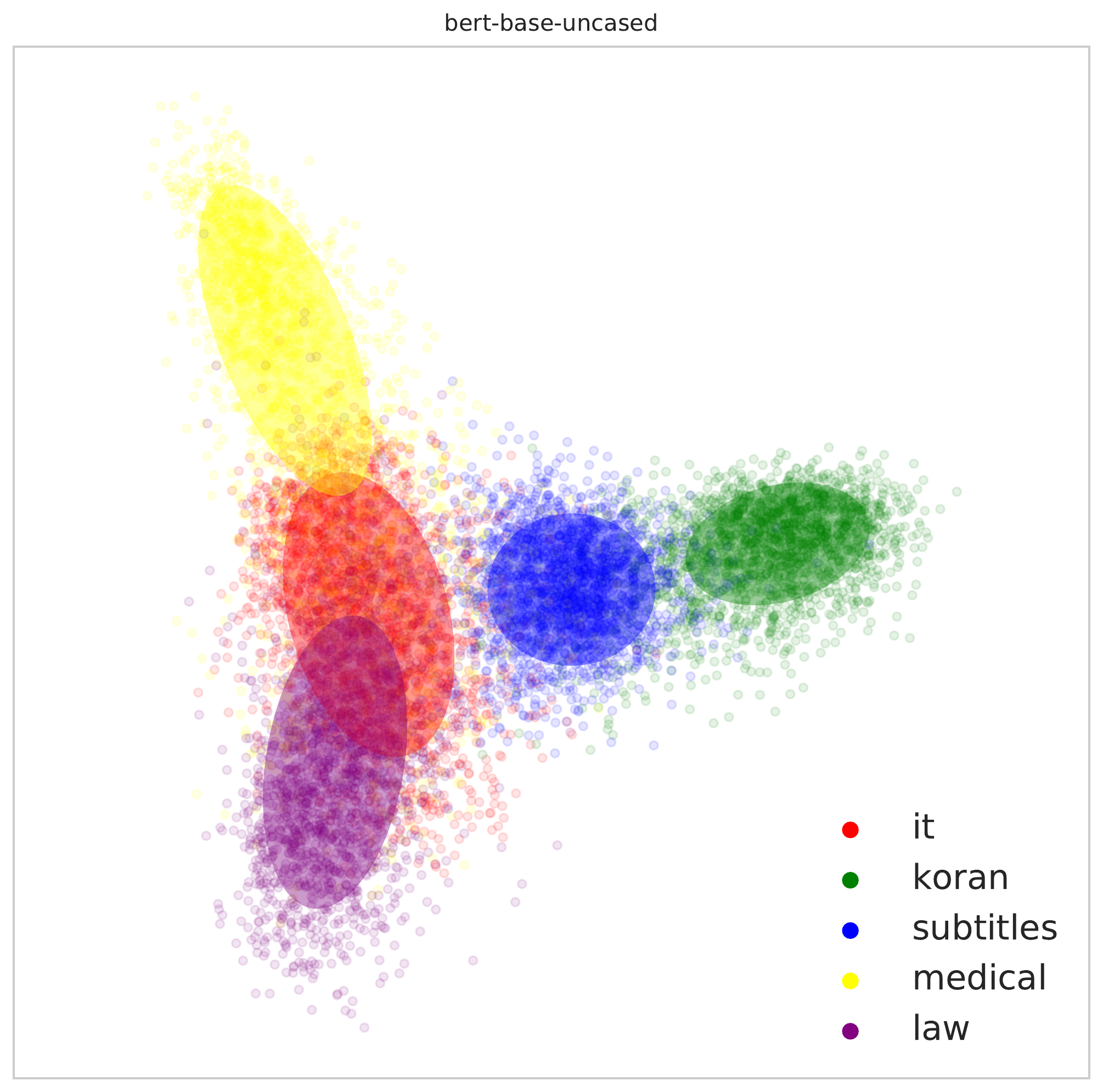}}
    \caption{A 2D visualization of the unsupervised GMM clustering for the same sentences as in Figure \ref{fig:bert_domains}.}
    \label{fig:bert_clusters}
\end{figure}

\subsection{Results and Discussion}
As can be seen in Table \ref{tab:cross_domain_accuracy}, pre-trained language models are indeed highly capable of generating sentence representations that cluster by domains, resulting in up to 87.66\%, 89.04\% and 89.94\% accuracy when using k=5, k=10 and k=15 clusters, respectively, across 10,000 sentences in 5 domains. We find these scores remarkably high given our straight-forward average-pooling strategy and that no domain-supervision was involved in the process of learning the pre-trained representations. Figure \ref{fig:bert_clusters} also demonstrates the quality of the obtained clusters in 2D using the BERT-base model, where the ellipses describe the mean and variance parameters learned for each cluster by the GMM with $k=5$.\footnote{Similar visualizations for additional models are available in the supplementary material.} 

We note that some classes of models did better than others: while all vector-based models did far better than the random and LDA baselines\footnote{Note that the LDA models were trained using the multi-domain data alone, and did not utilize additional pretraining as in the other, more successful models. This may explain their relatively weak performance.}, the MLM-based models dominated in all cases over word2vec and the auto-regressive models. This may be explained by the fact that the MLM-based models use the entire sentence context when generating the representations for each token, while the auto-regressive models only use the past context, and word2vec uses a limited window context. Using PCA improved performance in most cases and especially for the auto-regressive models, although the results for the MLMs remain high in both cases -- suggesting that these models encode the information very differently.

\begin{table*}[!t]
\vspace{-5px}
\begin{small}
\centering
\begin{tabular}{|l|l|}
\hline
\multicolumn{1}{|c|}{\textbf{Subtitles assigned to Koran}}                           & \multicolumn{1}{c|}{\textbf{Subtitles assigned to Medical}}                               \\ \hline
I am Spa'am, high priest of the boars.                             & Oxygen supply at 50\%.                                                  \\ \hline
Joseph, go in peace, and the Lord be with you.                     & Or it can help her walk again if the virus is kept in check             \\ 
                                                                   & with this.                                                              \\ \hline
\multicolumn{1}{|c|}{\textbf{Subtitles assigned to IT}}            & \multicolumn{1}{c|}{\textbf{Subtitles assigned to Law}}                                   \\ \hline
Push it up to the front of the screen.                             & Statutes, transcripts, redacted immunity agreements.                    \\ \hline
Polyalloy requires programming to take permanent                   & The Security Council therefore must press for his immediate             \\
form.                                                              &  referral to the International Criminal Court in The Hague.             \\ \hline
\multicolumn{1}{|c|}{\textbf{Law assigned to Medical}}             & \multicolumn{1}{c|}{\textbf{Law assigned to IT}}                           \\ \hline
- Viruses and virus-like organisms                                 &   "INFORMATION SOCIETY STATISTICS                                \\ \hline
where the glucose content is equal to or less than                 & This document must be attached to the certificate and field       \\
the fructose content.                                              & with it, except where there is a computerised checking system.     \\ \hline
\multicolumn{1}{|c|}{\textbf{Medical assigned to Law}}             & \multicolumn{1}{c|}{\textbf{Medical assigned to IT}}                           \\ \hline
This will be introduced by a Regulation adopted by the             & An updated and improved version of the CD-ROM was issued    \\  
European Commission.                                               &  to all subscribers during the first half of the year.              \\     \hline                  
The marketing authorisation was renewed on 22 May              & - All tables will be based on generic and not product-specific   \\
2002 and 22 May 2007.                                                   &  data.                                                         \\ \hline
\multicolumn{1}{|c|}{\textbf{IT assigned to Medical}}              & \multicolumn{1}{c|}{\textbf{IT assigned to Subtitles}}                           \\ \hline
R65: Harmful: may cause lung damage if swallowed                   & At the end we say good bye.                                 \\ \hline
Automatic Red-Eye Removal                                          & What would you like to do for your next shot?              \\ \hline

\end{tabular}
\caption{Sentences from one domain which were assigned to a cluster of another domain by the BERT-based clustering, k=5.}
\label{tab:error_analysis}
\end{small}
\vspace{-10px}
\end{table*}

\begin{figure}[!t]
    \centering
    \fcolorbox{white}{white}{\includegraphics[scale=0.75,trim={0cm 0cm 0cm 0.34cm},clip]{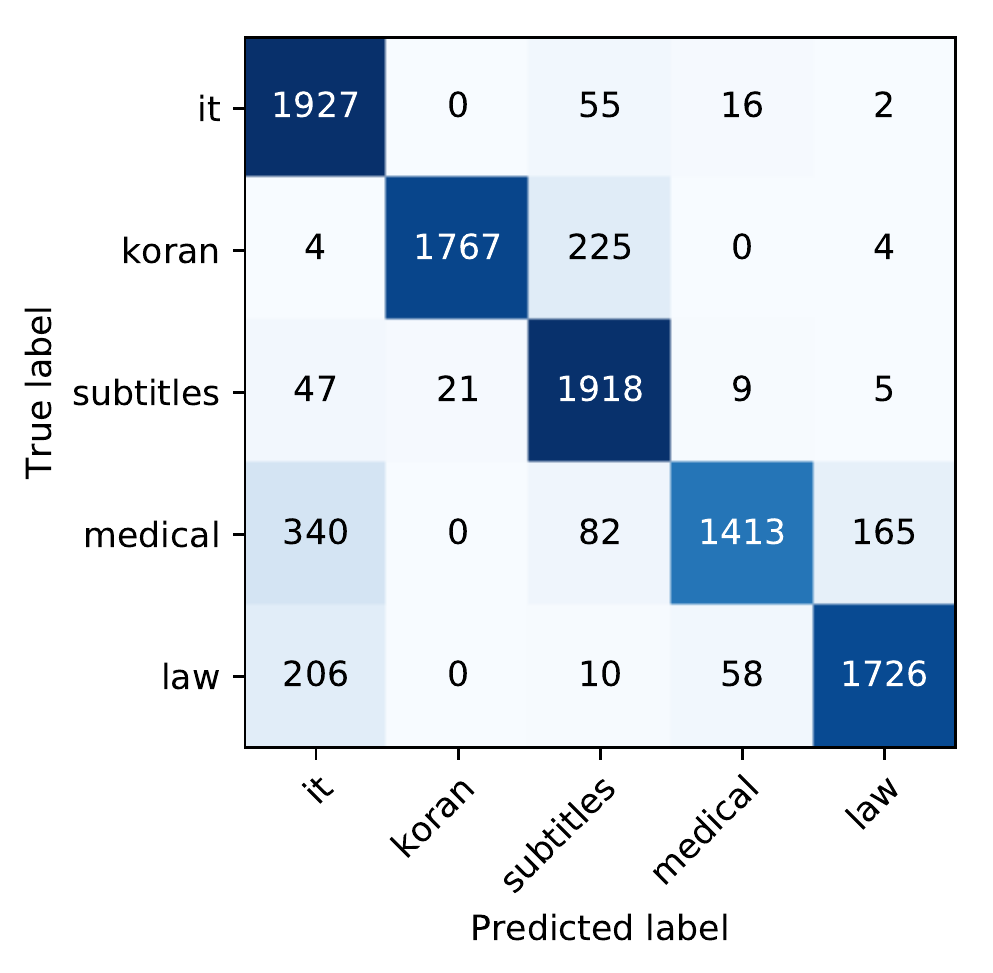}}
    \vspace{-20px}
    \caption{A confusion matrix for clustering with k=5 using BERT-base.}
    \label{fig:confusion}
    \vspace{-15px}
\end{figure}

\subsection{Analysis}
As can be seen in Figure \ref{fig:bert_clusters}, in some areas the domains are somewhat overlapping in the embedding space, which may lead to outlier cases where examples from one domain are assigned to a cluster of a another domain. We plot a confusion matrix (Figure \ref{fig:confusion}) to analyze this further based on the clustering with BERT-base and k=5. 
We first note that the outlier sentences are much shorter than the average sentence length in the corpus (11.62 tokens on average for outliers vs. 20.5 tokens on average in general). This makes sense as shorter sentences contain less information, making it harder to assign them to an appropriate cluster.
Table \ref{tab:error_analysis} shows examples of outlier sentences, assigned to clusters of domains different from their originating domain. We can see that in many cases the assignments are sensible -- for example for sentences originating from the subtitles corpus, a sentence that mentions ``great priest'' is assigned to the Koran cluster, a sentence that mentions ``The International Criminal Court in The Hague'' is assigned to the Law cluster, a sentence that mentions ``the virus'' is assigned to the Medical cluster and so on. This strengthens our claim that defining domains based on the corpus they originated from may be over-simplistic, and using a more data-driven approach may enable to find better domain assignments across different corpora.

The domain that attracted the largest number of outliers is the IT domain cluster, with 597 sentences assigned to it from other domains. Looking more closely we find that more than half of these sentences (340 out of 597) included numbers (e.g. ``34\% 25\% 34\%'' (from medical), ``(b) reference number 20 is deleted;'' (from law), ``(Command of Prostration \# 1)'' (from Koran) or ``The message, R2.'' (from subtitles)). As numbers appear in many different contexts, they may be harder to assign to a specific domain by the context-aware language models in such short sentences.
The second largest attractor of outliers is the Subtitles cluster, with 372 sentences assigned to it from other domains. We find that most of these sentences contain personal pronouns or question marks (228 out of 372, 61.2\%) while the ratio of such sentences in the entire corpus is only 40\%. Examples include ``Why did \textit{you} choose the name \& amarok;?'' (from IT), or ``What is Avonex?'' (from Medical). This may be expected as the subtitles corpus mainly includes transcriptions of spoken, conversational language, and ``conversation tends to have more verbs, more personal pronouns, and more questions'' \cite{conrad2005frequency}. Another possible reason for the subtitles domain to attract outliers is the fact that this is the least-topical cluster: movies and TV series may discuss diverse topics, unlike medical, religious, legal and technical texts that may have a more cohesive topic.

\section{Neural Machine Translation in a Multi-Domain Scenario}
As we showed that pre-trained language models are indeed very useful in clustering sentence representations by domains in an unsupervised manner, we now seek to harness this property for a down-stream task -- domain data selection for machine translation. Domain data selection is the task of selecting examples from a large corpus which are as close as possible to the domain of interest, given a smaller set of in-domain examples. 
The selected examples can be used to either (1) train a domain-specific model from scratch \cite{axelrod-etal-2011-domain}, (2) fine-tune a pre-trained general-domain model
\cite{sajjad2017neural,silva-etal-2018-extracting}, or (3) prioritize data for annotation as in an Active-Learning framework, if only monolingual data is available \cite{haffari-etal-2009-active}. 
To demonstrate the need for domain data selection and set the stage for our data selection experiments, we perform preliminary experiments with NMT in a multi-domain scenario.

\subsection{Multi-Domain Dataset}
\label{sec:dataset}
To simulate a diverse multi-domain setting we use the dataset proposed in \citet{koehn-knowles-2017-six}, as it was recently adopted for domain adaptation research in NMT \cite{hu-etal-2019-domain-adaptation, muller2019domain, dou2019unsupervised, dou-etal-2019-domain}.
The dataset includes parallel text in German and English from five diverse domains (Medical, Law, Koran, IT, Subtitles; as discussed in Section \ref{sec:clusters}), available via OPUS \cite{tiedemann-2012-parallel,aulamo-tiedemann-2019-opus}. 

In a preliminary analysis of the data we found that in both the original train/dev/test split by \citet{koehn-knowles-2017-six} and in the more recent split by \citet{muller2019domain} there was overlap between the training data and the dev/test data.\footnote{More details are available in the supplementary material.} 
Fixing these issues is important, as it may affect the conclusions one draws from experiments with this dataset. For example, as overlapping development sets favor memorization of the training set, one may choose checkpoints and report results on over-fitting models. This is especially relevant with neural sequence-to-sequence models, as they are highly susceptible to memorization \cite{aharoni-goldberg-2018-split} and hallucination \cite{Lee2018HallucinationsIN}, as confirmed by \citet{muller2019domain}.

\begin{table}[!t]
\centering
\begin{small}
\begin{tabular}{c|c|c|}
\cline{2-3}
                                & Original   & New Split  \\ \hline
\multicolumn{1}{|c|}{Medical}   & 1,104,752  & 248,099    \\ \hline
\multicolumn{1}{|c|}{Law}       & 715,372    & 467,309    \\ \hline
\multicolumn{1}{|c|}{IT}        & 378,477    & 222,927    \\ \hline
\multicolumn{1}{|c|}{Koran}     & 533,128    & 17,982     \\ \hline
\multicolumn{1}{|c|}{Subtitles} & 22,508,639 & 14,458,058 \\ \hline
\end{tabular}
\caption{Number of training examples for each domain in the original split \cite{muller2019domain} and in our split.}
\label{tab:new_split}
\end{small}
\vspace{-15px}
\end{table}

To create a better experimental setting to test generalization within and across domains, we create a new data split where we ensure that no such overlap between the training, development and test sets occur. We started from the split of \citet{muller2019domain} as it included newer versions of some of the datasets.\footnote{Their dataset is available in: \url{https://github.com/ZurichNLP/domain-robustness}} Furthermore, we did not allow more than one translation of a given source or target sentence, as such cases were very frequent in the dataset and usually stand for duplicate sentence pairs (See Table \ref{tab:new_split}). For example, applying this filtering reduced the size of the Koran corpus from 533,128 sentence pairs to only 17,982. Finally, following \citet{muller2019domain} we cap the subtitles corpus to 500,000 sentence pairs as it is much larger than the rest. We make the new split publicly available and hope it will enable better future experimentation on this important subject.\footnote{\url{https://github.com/roeeaharoni/unsupervised-domain-clusters}}

\subsection{Cross-Domain Experiments}
\textbf{Experimental Setup} We follow \citet{hu-etal-2019-domain-adaptation} and train domain-specific models for all domains. We then evaluate each model across the different domain test sets, enabling us to understand the effect of different domains on the downstream MT performance and to set up strong baselines for data selection experiments. We also train a general-domain model using the available data from all domains, as it is also a common approach in multi-domain scenarios \cite{muller2019domain}. In all experiments we use a similar Transformer \cite{aiayn_2017} model, and only control for the training data. More details on the exact training and hyperparameter settings for the NMT models are available in the supplementary material.

\textbf{Results} The results for the cross-domain evaluation are available in Table \ref{tab:cross_domain_bleu}. In most cases, the best results for each domain are obtained by training on the in-domain data. Training on all the available data helped mostly for the Koran test set. This is expected as the training data for this domain is considerably smaller than the training data for rest of the domains (Table \ref{tab:new_split}). We can also see that more data is not necessarily better \cite{gasco-etal-2012-data}: while the subtitles corpus is the largest of all 5 and includes 500,000 sentence pairs, it is second to last in performance as measured by the average BLEU across all test sets.

\begin{table}[!t]
\begin{small}
\begin{center}
\centering
\scalebox{0.98}{
\begin{tabular}{c|c|c|c|c|c|}
\cline{2-6}
                                & Medical       & Law         & Koran          & IT           & Subtitles     \\ \hline 
\multicolumn{1}{|c|}{Medical}   & \textbf{56.5} & 18.3        & 1.9            & 11.4         & 4.3           \\ \hline
\multicolumn{1}{|c|}{Law}       & 21.7          & \textbf{59} & 2.7            & 13.1        & 5.4           \\ \hline
\multicolumn{1}{|c|}{Koran}     & 0.1           & 0.2         & 15.9           & 0.2          & 0.5           \\ \hline
\multicolumn{1}{|c|}{IT}        & 14.9          & 9.6         & 2.8            & \textbf{43}  & 8.6           \\ \hline
\multicolumn{1}{|c|}{Subtitles} & 7.9           & 5.5         & 6.4            & 8.5          & 27.3          \\ \hline 
\multicolumn{1}{|c|}{All}       & 53.3          & 57.2        & \textbf{20.9}  & 42.1         & \textbf{27.6}  \\ \hline
\end{tabular}
}
\end{center}
\end{small}
\vspace{-10px}
\caption{SacreBLEU \cite{post-2018-call} scores of our baseline systems on the test sets of the new data split. Each row represents the results from one model on each test set. The best result in each column is marked in bold.}
\label{tab:cross_domain_bleu}
\vspace{-20pt}
\end{table}

\begin{figure}[!t]
    \centering
    \fcolorbox{white}{white}{\includegraphics[scale=0.57,trim={0.7cm 0.2cm 1.4cm 0.9cm},clip]{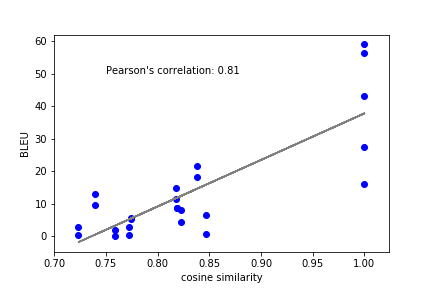}}
    \vspace{-15px}
    \caption{The cosine similarity between the centroids of the BERT representations for each domain pair vs. the corresponding cross-domain BLEU.}
    \label{fig:correlation}
    \vspace{-15px}
\end{figure}

\textbf{Cross-Domain BLEU vs. Cluster Proximity} An interesting observation can be made with respect to the visual analysis of the domain clusters as depicted in Figure \ref{fig:bert_clusters}: as the Medical cluster (in Yellow), Law cluster (in Purple) and IT cluster (in Red) are close to each other in the embedding space, their cross-domain BLEU scores are also higher. For example, note how in the results for the Medical domain-specific model (first row in Table \ref{tab:cross_domain_bleu}), the BLEU scores on the Law and IT test sets are much higher in comparison to those on the Koran and Subtitles test sets, which clusters are farther away in the visualized embedding space. Similarly, as the Subtitles cluster (Blue) is closer to the Koran cluster (Green), the highest cross-domain BLEU score on the Koran test set is from the Subtitles model. To further quantify this phenomenon, we plot and measure Pearson's correlation between the cosine similarity of the centroids for the English BERT-based dev sentence representations for each domain pair, and the cross-domain BLEU score for this domain pair. This is shown in Figure \ref{fig:correlation}. We can see the general trend where the closer the domain centroids are (with a similarity of 1 for training and evaluating on the same domain), the higher the cross-domain BLEU is between those domains, resulting in a Pearson's correlation of 0.81 (strong correlation). This suggests that such preliminary visual analysis can be a useful tool for understanding the relationship between diverse datasets, and motivates the use of pre-trained language model representations for domain data selection in MT.

\section{Domain Data Selection with Pretrained Language Models}
As shown in the previous section, using the right data is critical for achieving good performance on an in-domain test set, and more data is not necessarily better. However, in real-world scenarios, the availability of data labeled by domain is limited, e.g. when working with large scale, web-crawled data. In this section we focus on a data-selection scenario where only a very small number of in-domain sentences are used to select data from a larger unlabeled parallel corpus. An established method for data selection was proposed by \citet{moore-lewis-2010-intelligent}, which was also used in training the winning systems in WMT 2019 \cite{ng-etal-2019-facebook, barrault-etal-2019-findings}. This method compares the cross-entropy, according to domain-specific and non-domain-specific language models, for each candidate sentence for selection. The sentences are then ranked by the cross-entropy difference, and only the top sentences are selected for training. 

While the method by \citet{moore-lewis-2010-intelligent} is tried-and-true, it is based on simple n-gram language models which cannot generalize beyond the n-grams that are seen in the in-domain set. 
In addition, it is restricted to the in-domain and general-domain datasets it is trained on, which are usually small. On the contrary, pre-trained language models are trained on massive amounts of text, and, as we showed through unsupervised clustering, learn representations with domain-relevant information. In the following sections, we investigate whether this property of pretrained language models makes them useful for domain data selection.

\subsection{Methods}
We propose two methods for domain data selection with pretrained language models.

\textbf{Domain-Cosine} In this method we first compute a query vector, which is the element-wise average over the vector representations of the sentences in the small in-domain set. We use the same sentence-level average-pooling approach as described in Section \ref{sec:clusters} to obtain sentence representations. We then retrieve the most relevant sentences in the training set by computing the cosine similarity of each sentence with this query vector and ranking the sentences accordingly.

\begin{table}[!b]
\vspace{-10px}
\centering
\begin{small}
\scalebox{0.92}{
\begin{tabular}{c|c|c|c|c|c|c|}
\cline{2-7}
                                & \multicolumn{3}{c|}{without pre-ranking} & \multicolumn{3}{c|}{with pre-ranking} \\ \cline{2-7} 
                                & p         & r         & F1        & p         & r        & F1       \\ \hline
\multicolumn{1}{|c|}{Subtitles} & 0.722     & 0.984     & 0.833     & 0.964     & 0.978    & 0.971    \\ \hline
\multicolumn{1}{|c|}{Law}       & 0.761     & 0.94      & 0.841     & 0.944     & 0.94     & 0.942    \\ \hline
\multicolumn{1}{|c|}{Medical}   & 0.821     & 0.916     & 0.866     & 0.929     & 0.92     & 0.925    \\ \hline
\multicolumn{1}{|c|}{IT}        & 0.848     & 0.956     & 0.898     & 0.955     & 0.98     & 0.967    \\ \hline
\multicolumn{1}{|c|}{Koran}     & 0.966     & 0.958     & 0.962     & 0.994     & 0.974    & 0.984    \\ \hline
\end{tabular}
}
\end{small}
\vspace{-5px}
\caption{Ablation analysis showing precision (p) recall (r) and F1 for the binary classification accuracy on a held-out set, with and without pre-ranking.}
\label{tab:preranking_ablation}
\vspace{-5px}
\end{table}

\begin{table*}[ht!]
\begin{small}
\centering
\begin{tabular}{l|c|c|c|c|c|c|}
\cline{2-7}
                                                              & Medical        & Law         & Koran       & IT             & Subtitles          & Average \\ \cline{1-7} 
\multicolumn{1}{|l|}{Random-500k}                              & 49.8          & 53.3        & 18.5        & 37.5           & 25.5               &   36.92 \\ \hline
\multicolumn{1}{|l|}{Moore-Lewis-Top-500k}                     & 55            & 58          & 21.4        & 42.7           & 27.3               &   40.88 \\ \hline
\multicolumn{1}{|l|}{Domain-Cosine-Top-500k}                   & 52.7          & 58          & \textbf{22} & 42.5           & 27.1               &   40.46 \\ \hline
\multicolumn{1}{|l|}{Domain-Finetune-Top-500k}                 & 54.8          & 58.8        & 21.8        & \textbf{43.5}  & 27.4               &   \textbf{41.26} \\ \hline
\multicolumn{1}{|l|}{Domain-Finetune-Positive}                 & 55.3          & 58.7        & 19.2        & 42.5           & 27                 &   40.54 \\ \hline
\multicolumn{1}{|l|}{Oracle}                                   & \textbf{56.5} & \textbf{59} & 15.9        & 43             & 27.3               &   40.34 \\ \hline
\multicolumn{1}{|l|}{All}                                      & 53.3          & 57.2        & 20.9        & 42.1           & \textbf{27.6}      &   40.22 \\ \hline
\end{tabular}
\vspace{-5px}
\caption{SacreBLEU scores for the data selection experiments. Highest scores per column are marked in bold.}
\label{tab:bleu_selection}
\end{small}
\vspace{-10px}
\end{table*}

\textbf{Domain-Finetune} It is now common knowledge that pretrained language models are especially useful when fine-tuned for the task of interest in an end-to-end manner \cite{ruder-etal-2019-transfer}. In this method we fine-tune the pretrained LM for binary classification, where we use the in-domain sentences as positive examples, and randomly sampled general-domain sentences as negative examples. We then apply this classifier on the general-domain data and pick the sentences that are classified as positive as in-domain, or choose the top-k sentences as ranked by the classifier output distribution. This can be seen as an instance of positive-unlabeled learning for document-set expansion; see \citet{jacovi2019scalable} for a recent discussion and methodology for this task.

\textbf{Negative Sampling with Pre-ranking} One problem that may rise when randomly sampling negative examples is that unlabeled in-domain sentences from the general-domain data may be sampled as negative examples -- deteriorating the classifier performance. To alleviate this issue, we perform a biased sampling of negative examples. We first rank the general-domain data using the Domain-Cosine method, and then sample negative examples under a certain threshold in the ranking (in our experiments we sampled from the bottom two-thirds). Table \ref{tab:preranking_ablation} shows an ablation for such pre-ranking, measuring precision, recall and F1 for binary classification on a held-out set for each domain. When not using pre-ranking, as the training data for the domain is larger, the precision is lower -- since more in-domain examples are drawn as negative samples. Using pre-ranking indeed alleviates this issue, achieving higher F1 scores in all cases. Given the results in Table \ref{tab:preranking_ablation} we always use pre-ranking in the following experiments.

\subsection{Experimental Setup}
We perform data selection experiments for each domain in the multi-domain dataset. As the small set of monolingual in-domain data we take the 2000 development sentences from each domain. For the general-domain corpus we concatenate the training data from all domains, resulting in 1,456,317 sentences. To enable faster experimentation we used DistilBERT \cite{sanh2019distilbert} for the Domain-Cosine and Domain-Finetune methods. More technical details are available in the supplementary material. We compare our methods to four approches: (1) The established method by \citet{moore-lewis-2010-intelligent}, (2) a random selection baseline, (3) an oracle which is trained on all the available in-domain data, and (4) the model we train on all the domains concatenated. We select the top 500k examples to cover the size of every specific in-domain dataset. We train Transformer NMT models on the selected data with a similar configuration to the ones trained in the cross-domain evaluation. 

\begin{table}[!t]
\begin{small}
\centering
\scalebox{0.9}{
\begin{tabular}{c|c|c|c|c|c|c|}
\cline{2-7}
                                & \multicolumn{2}{c|}{Moore-Lewis} & \multicolumn{2}{c|}{D-Cosine} & \multicolumn{2}{c|}{D-Finetune} \\ \cline{2-7} 
                                & p               & r              & p                & r               & p                 & r                \\ \hline
\multicolumn{1}{|c|}{Medical}   & 0.476            & 0.955           & 0.391             & 0.788            & 0.485              & 0.975             \\ \hline
\multicolumn{1}{|c|}{Law}       & 0.836            & 0.894           & 0.841             & 0.899            & 0.902              & 0.965             \\ \hline
\multicolumn{1}{|c|}{Koran}     & 0.35             & 0.985           & 0.36              & 0.989            & 0.36               & 0.998             \\ \hline
\multicolumn{1}{|c|}{IT}        & 0.441            & 0.985           & 0.382             & 0.857            & 0.447              & 0.998             \\ \hline
\multicolumn{1}{|c|}{Subtitles} & 0.899            & 0.899           & 0.916             & 0.916            & 0.957              & 0.957             \\ \hline
\multicolumn{1}{|c|}{Average}   & 0.6              & 0.944           & 0.578             & 0.89             & 0.63               & 0.979            \\ \hline
\end{tabular}
}
\vspace{-5px}
\caption{Precision (p) and recall (r) for data selection of 500k sentences with respect to the oracle selection.}
\label{tab:selection_analysis}
\end{small}
\vspace{-15px}
\end{table}

\subsection{Results}
The results are available in Table \ref{tab:bleu_selection}. We can see that all selection methods performed much better in terms of BLEU than random selection. It is also nice to see that all selection methods performed better than using all the available data or the oracle-selected data when averaged across all domains, showing again that more data is not necessarily better in multi-domain scenarios and that data selection is a useful approach. Regarding a comparison of the data selection methods, Moore-Lewis performed better than Domain-Cosine, while Domain-Finetune performed best, showing the benefit of fine-tuning large pretrained models for the data selection task. Using the positively-labeled examples alone (Domain-Finetune-Positive) performed worse than using the top 500k examples but better than Domain-Cosine, while not requiring to determine the number of selected sentences.  
\subsection{Analysis}
We perform an analysis on the selected datasets, where we measure the precision and recall of sentence selection with respect to the oracle selection. The results are available in Table \ref{tab:selection_analysis}. As also reflected in the BLEU scores, the Domain-Finetune method resulted in the highest domain recall with a minimum of 97.5, while Moore-Lewis and Domain-Cosine scored 89.4 and 78.8 respectively. We find these results very appealing given that only 2000 in-domain sentences were used for selection for each domain out of 1.45 million sentences. Also note that we used DistilBERT in these experiments: we believe that using larger, non-distilled models may result in even better selection performance (although at the price of larger computational requirements).

\vspace{-5px}
\section{Related Work}
\vspace{-5px}
Previous works used n-gram LMs for data selection \cite{moore-lewis-2010-intelligent,axelrod-etal-2011-domain} or other count-based methods \cite{axelrod2017cynical,poncelas2018data,parcheta2018data,santamaria2019data}. While such methods work well in practice, they cannot generalize beyond the N-grams observed in the in-domain datasets, which are usually small.

\citet{duh-etal-2013-adaptation} proposed to replace n-gram models with RNN-based LMs with notable improvements. However, such methods do not capture the rich sentence-level global context as in the recent self-attention-based MLMs; as we showed in the clustering experiments, autoregressive neural LMs were inferior to masked LMs in clustering the data by domain. In addition, training very large neural LMs may be prohibitive without relying on pre-training. 

Regarding domain clustering for MT, \citet{hasler-etal-2014-dynamic-topic} discovered topics using LDA instead of using domain labels. \citet{cuong-etal-2016-adapting} induced latent subdomains from the training data using a dedicated probabilistic model.

Many works used vector-based retrieval for data selection; \citet{ruder-plank-2017-learning} learn to select data using Bayesian optimization, and explored word2vec for that purpose. \citet{duma-menzel-2016-data} create paragraph vectors for data selection in the context of SMT. \citet{wang-etal-2017-sentence} use internal representations from the NMT model to perform data selection. \citet{bapna-firat-2019-non} propose a mechanism for incorporating retrieved sentences for each instance for domain adaptation in NMT, using representations extracted from a pre-trained NMT model. \citet{farajian2017multi} explored instance-based data selection in a multi-domain scenario using information retrieval methods. 

Other related works on domain adaptation include \citet{dou2019unsupervised} that adapts multi-domain NMT models with domain-aware feature embeddings, which are learned via an auxiliary language modeling task. \citet{peris2017neural} proposed neural-network based classifiers for data selection in SMT. For more related work on data selection and domain adaptation in the context of MT, see the surveys by \citet{eetemadi2015survey} for SMT and more recently \citet{chu-wang-2018-survey} for NMT. 

Unrelated to MT, \citet{ma-etal-2019-domain} used BERT to select data for tasks from the GLUE benchmark \cite{wang-etal-2018-glue}. However, they assumed supervision for all the different tasks/domains, while we propose an unsupervised method requiring only a small set of in-domain data. Also in the context of pretrained language models, \citet{gururangan2020dont} show the importance of additional pretraining with in-domain data to improve the down-stream task-specific performance.

While previous work made important contributions to domain data selection, our work is the first to explore massive pretrained language models for both unsupervised domain clustering and for data selection in NMT.

\section{Conclusions and Future Work}
We showed that massive pre-trained language models are highly effective in mapping data to domains in a fully-unsupervised manner using average-pooled sentence representations and GMM-based clustering. We suggest that such clusters are a more appropriate, data driven approach to domains in natural language than simplistic labels (e.g. ``medical text''), and that it will improve over time as better and larger pretrained LMs will become available. We proposed new methods to harness this property for domain data selection using distance-based ranking in vector space and pretrained LM fine-tuning, requiring only a small set of in-domain data. We demonstrated the effectiveness of our methods on a new, improved data split we created for a previously studied multi-domain machine translation benchmark. Our methods perform similarly or better than an established data selection method and oracle in-domain training across all five domains in the benchmark. 

This work just scratches the surface with what can be done on the subject; possible avenues for future work include extending this with multilingual data selection and multilingual LMs \cite{conneau2019cross, conneau2019unsupervised,wu2019emerging,hu2020xtreme}, using such selection methods with domain-curriculum training \cite{zhang-etal-2019-curriculum,wang-etal-2019-dynamically}, applying them on noisy, web-crawled data \cite{junczys-dowmunt-2018-dual} or for additional tasks \cite{gururangan2020dont}. Another interesting avenue is applying this to unsupervised NMT, which is highly sensitive to domain mismatch \cite{marchisio2020does,kim2020and}. We hope this work will encourage more research on finding the right data for the task, towards more efficient and robust NLP.

\section*{Acknowledgements}
We thank Wei Wang for early discussions on domain adaptation and data selection that inspired this work during Roee's internship in Google Translate.

\bibliography{anthology,domains}
\bibliographystyle{acl_natbib}

\clearpage
\appendix
\section{Appendix}
\label{sec:appendix}
\subsection{NMT Training} Figure \ref{fig:fairseq_config} details the hyperparameter configuration we used to train the NMT models. We use Transformer models \cite{aiayn_2017} in the Base configuration using the implementation provided in Fairseq \cite{ott-etal-2019-fairseq}.  For all models we use a joint BPE vocabulary \cite{sennrich-etal-2016-neural} learned with 32k merge operations over the concatenated corpus in both languages, enabling to tie all the embedding layers \cite{press-wolf-2017-using}.\footnote{We used the implementation in \url{https://github.com/rsennrich/subword-nmt}} We perform early stopping if the BLEU score on the domain-specific development set did not improve in 10 consequent checkpoints. We use the ADAM \cite{kingma2014adam} optimizer with an initial learning rate of $5\cdot{}10^-4$ and a maximum of 4096 tokens per batch. We trained all models on a single NVIDIA GPU. We decode using beam search with a beam size of 5. For pre-processing we used the Moses \cite{koehn-etal-2007-moses} pipeline including tokenization, normalize-punctuation, non-printing character removal, truecasing and cleaning. We removed examples with sequences longer than 100 tokens from the training data (before subword segmentation). 

\subsection{Data Split} Table \ref{tab:data_splits} shows details about the overlap between the training, development and test sets for the different data splits of the multi-domain dataset. The overlap was computed using the English part of the corpus.

\subsection{GMM Clustering} We learn GMMs with full covariance matrices, i.e. without constraints on covariance matrices that determine the shape of each component in the mixture, as implemented in scikit-learn \cite{scikit-learn}. We train the models until convergence or for a maximum of 150 EM iterations. 

\subsection{Language Model Finetuning} We fine-tune the binary classification head for 5 epochs. We use the ADAM \cite{kingma2014adam} optimizer with an initial learning rate of $2\cdot{}10^-5$. We train the model using 4 NVIDIA GPUs with 256 sentences per batch (64 per GPU).

\subsection{Moore-Lewis Implementation} We used the implementation of \citet{moore-lewis-2010-intelligent} by Pamela Shapiro, as available in: \url{https://github.com/pamelashapiro/moore-lewis}. This implementation uses the KenLM N-Gram language model toolkit \cite{heafield-2011-kenlm}. 

\subsection{Additional Visualizations} 
\begin{figure*}[t]
    \centering
    \fcolorbox{white}{white}{\includegraphics[scale=0.5,trim={0.0cm 0.0cm 0.0cm 0.0cm},clip]{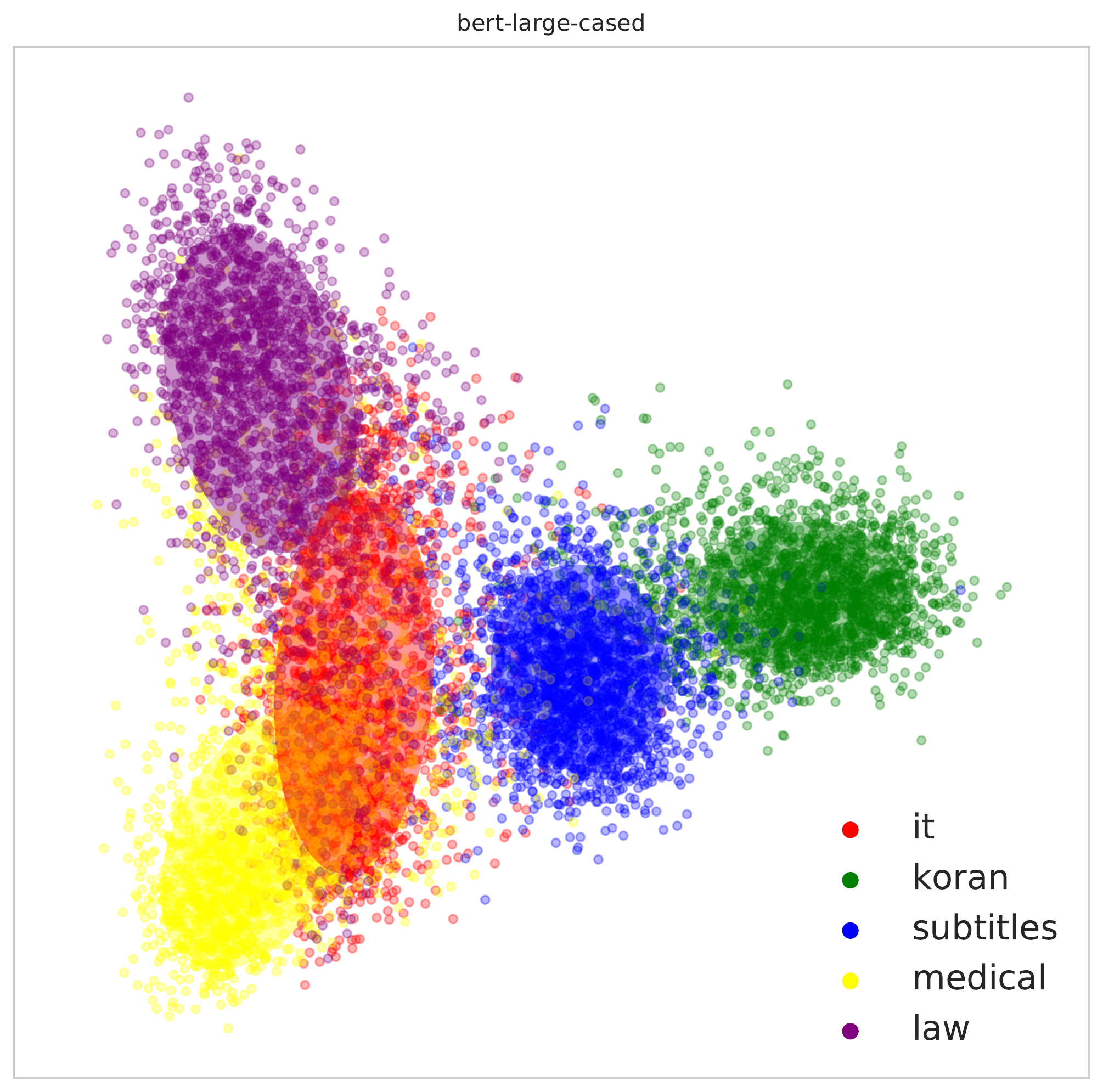}}
    \fcolorbox{white}{white}{\includegraphics[scale=0.5,trim={0.0cm 0.0cm 0.0cm 0.0cm},clip]{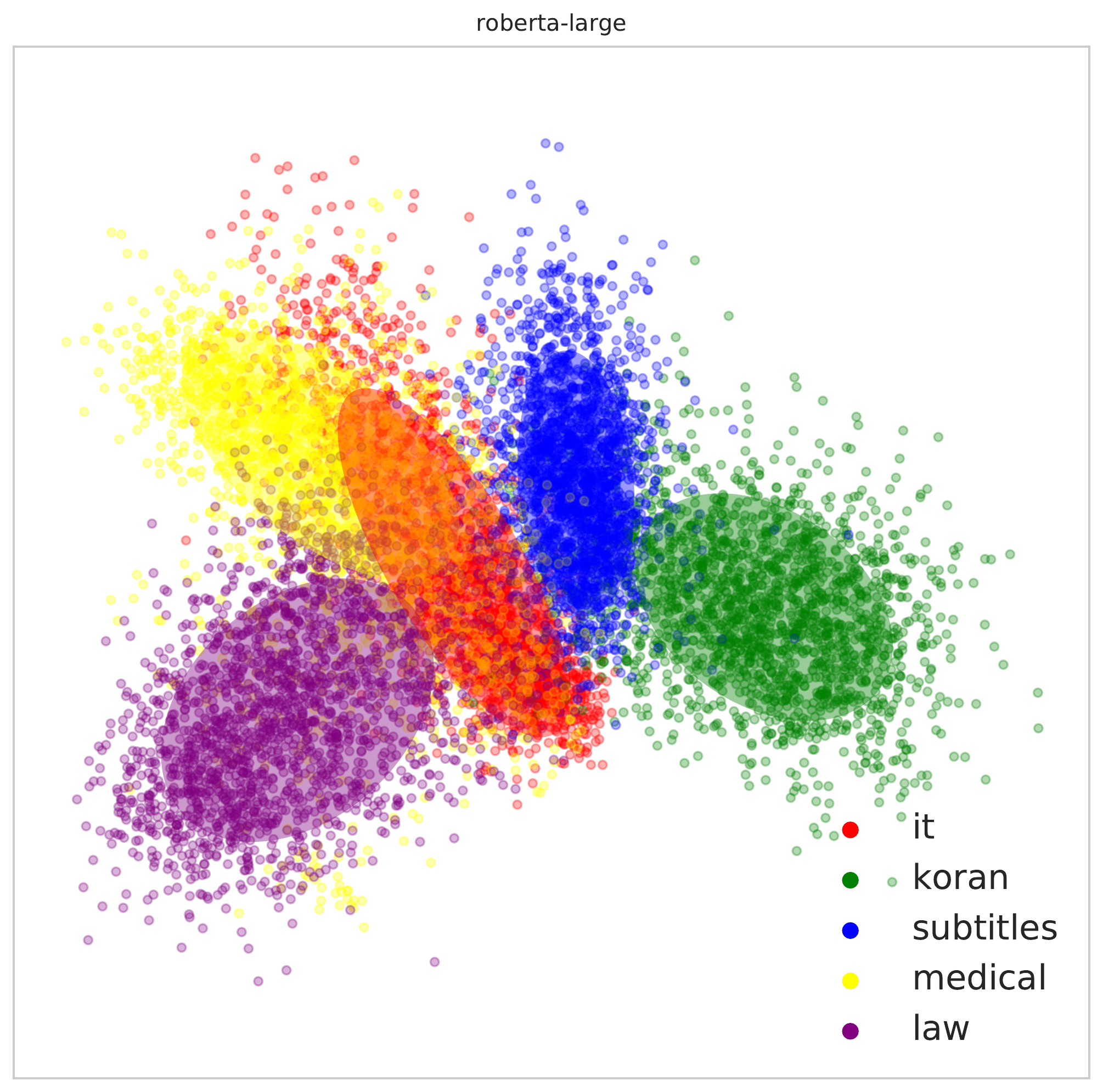}}
    \caption{2D visualizations of the unsupervised GMM-based clustering for different pretrained MLMs.}
    \label{fig:more_viz_mlm}
    \vspace{-15px}
\end{figure*}

\begin{figure*}[t]
    \centering
    \fcolorbox{white}{white}{\includegraphics[scale=0.5,trim={0.0cm 0.0cm 0.0cm 0.0cm},clip]{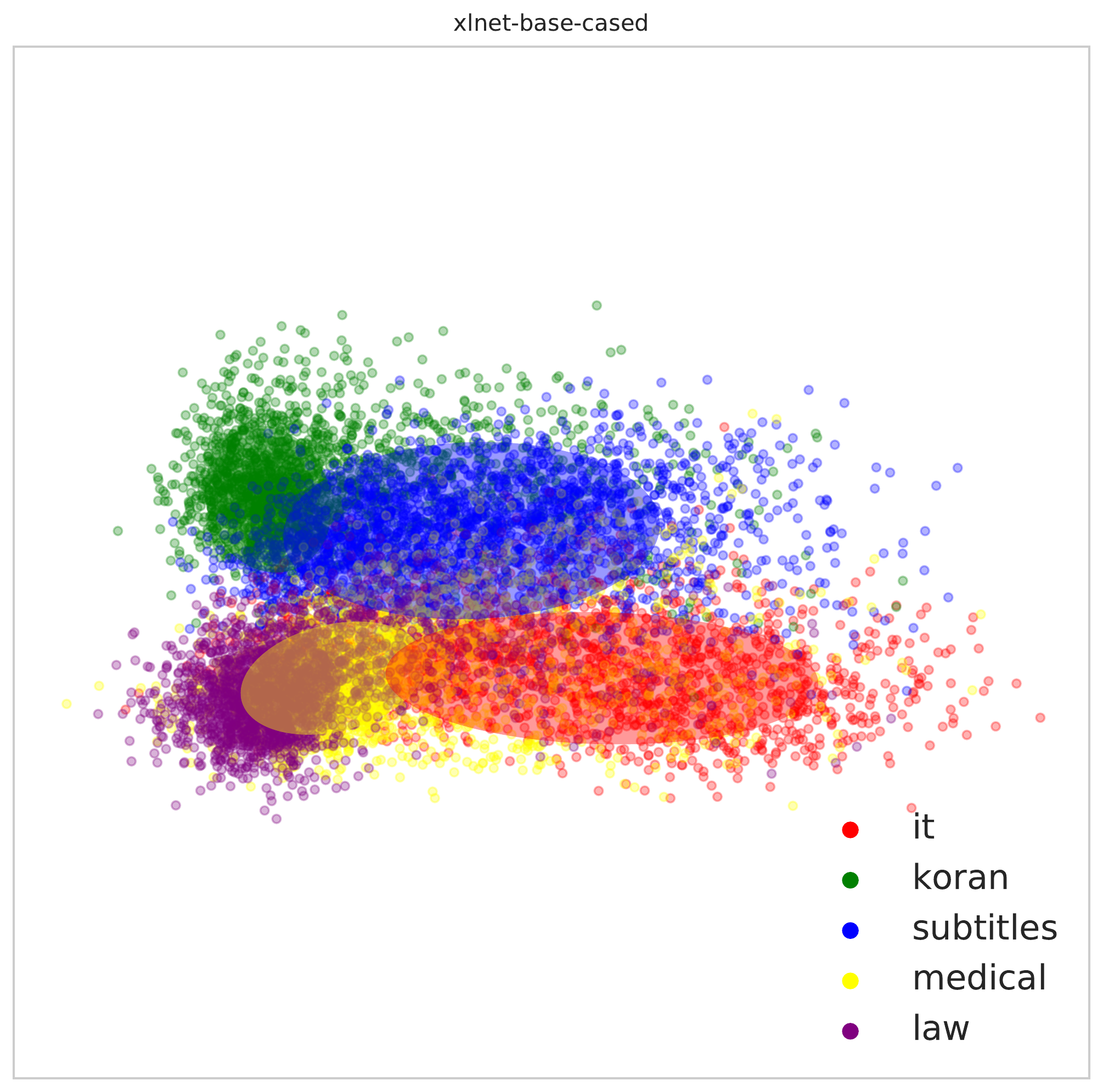}}
    \fcolorbox{white}{white}{\includegraphics[scale=0.5,trim={0.0cm 0.0cm 0.0cm 0.0cm},clip]{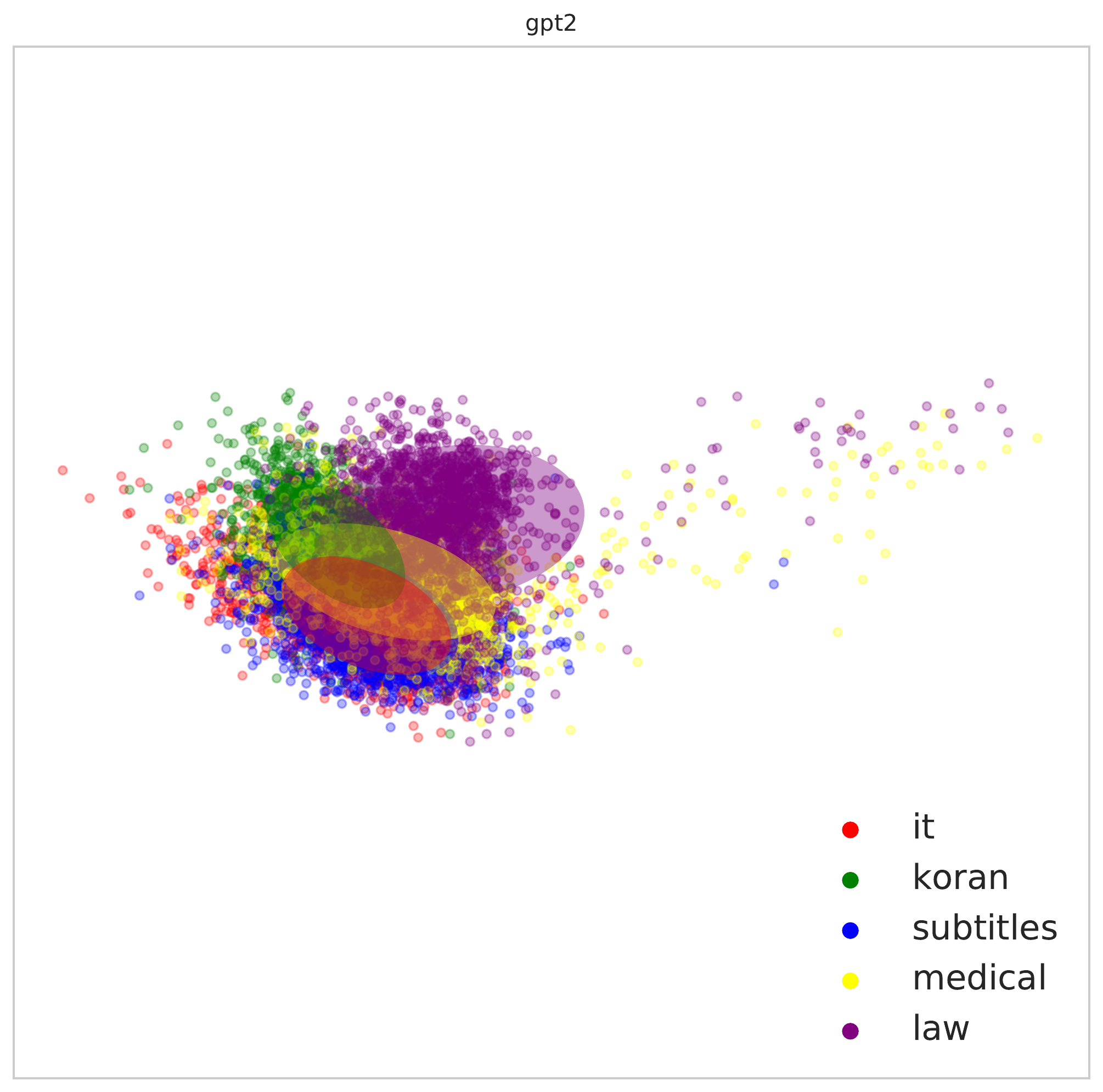}}
    \caption{2D visualizations of the unsupervised GMM-based clustering for different pretrained auto-regressive LMs.}
    \label{fig:more_viz_auto}
    \vspace{-15px}
\end{figure*}

Figure \ref{fig:more_viz_mlm} shows visualizations of the multi-domain dataset from additional pre-trained masked language models (BERT large and RoBERTa), and Figure \ref{fig:more_viz_auto} shows the same visualization for autoregressive models (XLNet and GPT2).

\begin{table*}[!hb]
\begin{small}
\begin{center}
\begin{tabular}{cc|c|c|c|}
\cline{3-5}
                                                                                                         &           & \citet{koehn-knowles-2017-six}      & \citet{muller2019domain}         & New Split \\ \hline
\multicolumn{1}{|c|}{\multirow{5}{*}{\begin{tabular}[c]{@{}c@{}}\% dev \\ in train\end{tabular}}}        & Medical   &  1090/2000 (54.5\%) &    1204/2000 (60.2\%) & 0/2000         \\ \cline{2-5} 
\multicolumn{1}{|c|}{}                                                                                   & Koran     &  0/2000   &  1926/2000 (96.3) & 0/2000         \\ \cline{2-5} 
\multicolumn{1}{|c|}{}                                                                                   & Subtitles &  1183/5000 (23.66\%)  &  638/2000 (31.9\%) & 0/2000         \\ \cline{2-5} 
\multicolumn{1}{|c|}{}                                                                                   & Law       &   595/2000 (29.75\%)   & 1000/2000 (50\%)  & 0/2000         \\ \cline{2-5} 
\multicolumn{1}{|c|}{}                                                                                   & IT        &  2496/2526 (98.81\%)   &  783/2000 (39.15\%)  & 0/2000         \\ \hline
\multicolumn{1}{|c|}{\multirow{5}{*}{\begin{tabular}[c]{@{}c@{}}\% test \\ in train\end{tabular}}}       & Medical   &   571/2000 (28.55\%)   & 516/1691 (30.51\%) & 0/2000         \\ \cline{2-5} 
\multicolumn{1}{|c|}{}                                                                                   & Koran     &      0/2000          &  1949/2000 (97.45\%)  & 0/2000         \\ \cline{2-5} 
\multicolumn{1}{|c|}{}                                                                                   & Subtitles &   451/5000 (9.02\%)  &  478/2000 (23.9\%) & 0/2000         \\ \cline{2-5} 
\multicolumn{1}{|c|}{}                                                                                   & Law       &  649/2000 (32.45\%)  &  966/2000 (48.3\%)    & 0/2000         \\ \cline{2-5} 
\multicolumn{1}{|c|}{}                                                                                   & IT        &      945/1856 (50.92\%)    & 1036/2000 (51.8\%) & 0/2000         \\ \hline

\end{tabular}
\caption{Details about the different data splits for the multi-domain corpus.}
\label{tab:data_splits}
\end{center}
\end{small}
\end{table*}

\begin{figure}[!t]
    \begin{center}
    \begin{tiny}
    \begin{verbatim}
CUDA_VISIBLE_DEVICES=0 \
python $FAIRSEQ_PATH/train.py ${BINARIZED_DATA_DIR} \
  --arch transformer_wmt_en_de \
  --share-all-embeddings \
  --optimizer adam \
  --adam-betas '(0.9, 0.98)' \
  --clip-norm 1.0 \
  --lr 0.0005 \
  --lr-scheduler inverse_sqrt \
  --warmup-updates 4000 \
  --warmup-init-lr 1e-07 \
  --dropout 0.2 \
  --weight-decay 0.0 \
  --criterion label_smoothed_cross_entropy \
  --label-smoothing 0.1 \
  --max-tokens 4096 \
  --update-freq 5 \
  --attention-dropout 0.2 \
  --activation-dropout 0.2 \
  --max-epoch 200 \
  --seed 17 \
  -s $src \
  -t $tgt \
  --save-dir $MODEL_PATH \
  --save-interval-updates 10000 \
  --validate-interval 1      
  \end{verbatim}
  \end{tiny}
  \end{center}
  \vspace{-20px}
    \caption{The hyperparameter configuration we used for NMT model training using Fairseq \cite{ott-etal-2019-fairseq}.}
    \label{fig:fairseq_config}
\end{figure}

\end{document}